\documentclass[10pt,twocolumn,letterpaper]{article}

\usepackage{cvpr}              
\definecolor{cvprblue}{rgb}{0.21,0.49,0.74}
\usepackage[pagebackref,breaklinks,colorlinks,allcolors=cvprblue]{hyperref}
\usepackage{multirow}
\usepackage{amsmath}
\usepackage{amsfonts}
\usepackage{xcolor,colortbl}
\usepackage{booktabs}
\usepackage{arydshln}
\usepackage{makecell} 
\usepackage{amssymb}
\usepackage{afterpage}

\usepackage[title]{appendix}

\def\paperID{*****} 
\def\confName{CVPR}
\def\confYear{2026}

\title{No Need For Real Anomaly: \\MLLM Empowered Zero-Shot Video Anomaly Detection}

\author{
    Zunkai Dai$^1$, \quad Ke Li$^{1, \dagger}$, \quad Jiajia Liu$^2$, \quad Jie Yang$^1$, \quad Yuanyuan Qiao$^{1, \dagger}$\\
    $^1$Beijing University of Posts and Telecommunications \\
    $^2$Northwestern Polytechnical University \\
    {\tt\small \{daizk, like1990, janeyang, yyqiao\}@bupt.edu.cn, liujiajia@nwpu.edu.cn}
}

\begin{document}

\twocolumn[{
\renewcommand\twocolumn[1][]{#1}
\maketitle
\begin{center}
    \captionsetup{type=figure}
    \includegraphics[width=.99\textwidth]{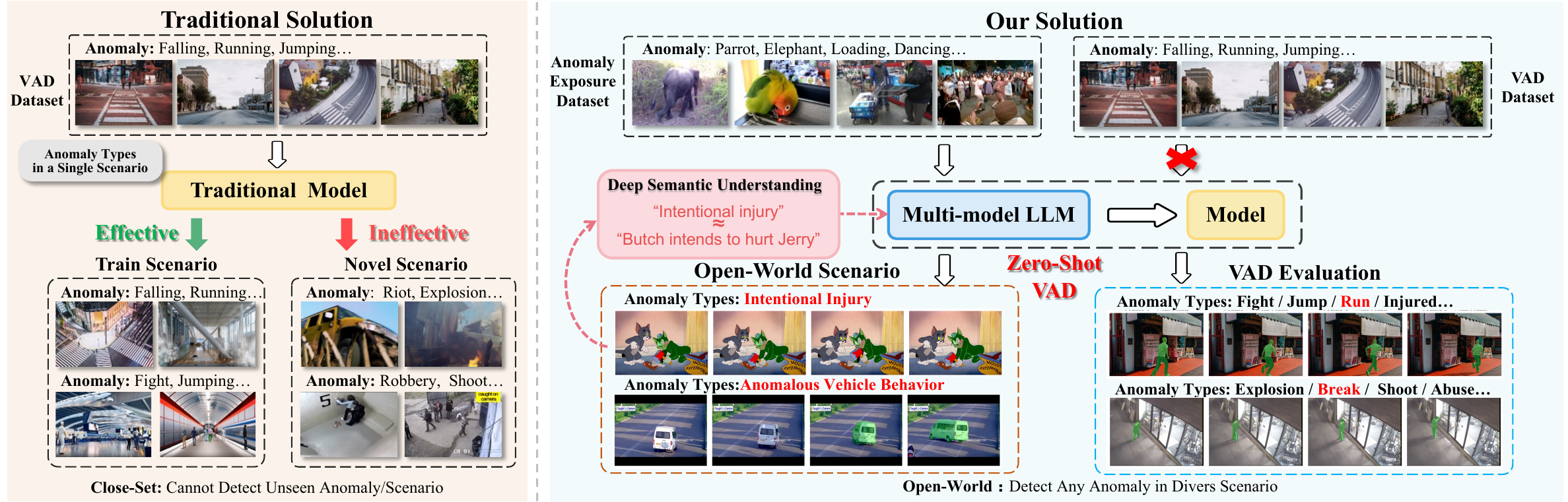}
    \captionof{figure}{\textbf{Motivation}. \textbf{Left:} Existing VAD methods rely on training with anomaly data from single scenarios, resulting in poor generalization capability to novel anomaly types or unseen scenarios. \textbf{Right:} Our LAVIDA model leverages MLLM to understand deep anomaly semantics, enabling generalization to arbitrary anomaly types across diverse scenarios. The training data consists of pseudo anomaly data synthesized from external datasets, without incorporating any VAD data.}
\end{center}
}]

\let\thefootnote\relax\footnotetext{$^\dagger$Corresponding author.}

\begin{abstract}
The collection and detection of video anomaly data has long been a challenging problem due to its rare occurrence and spatio-temporal scarcity. Existing video anomaly detection (VAD) methods under perform in open-world scenarios. Key contributing factors include limited dataset diversity, and inadequate understanding of context-dependent anomalous semantics. To address these issues, i) we propose LAVIDA, an end-to-end zero-shot video anomaly detection framework. ii) LAVIDA employs an Anomaly Exposure Sampler that transforms segmented objects into pseudo-anomalies to enhance model adaptability to unseen anomaly categories. It further integrates a Multimodal Large Language Model (MLLM) to bolster semantic comprehension capabilities. Additionally, iii) we design a token compression approach based on reverse attention to handle the spatio-temporal scarcity of anomalous patterns and decrease computational cost. The training process is conducted solely on pseudo anomalies without any VAD data. Evaluations across four benchmark VAD datasets demonstrate that LAVIDA achieves SOTA performance in both frame-level and pixel-level anomaly detection under the zero-shot setting. Our code is available in \url{https://github.com/VitaminCreed/LAVIDA}. 
\end{abstract}    
\section{Introduction}
\label{sec:intro}
Video Anomaly Detection (VAD) aims to identify behaviors that deviate from normal patterns or represent unexpected events in video sequences. Classical VAD methods assume static scenes, closed-set anomaly categories, and stationary data distributions, yet these assumptions fail in dynamic environments where scenes evolve, behaviors shift, and distributions drift continuously. Recent works have reconceptualized VAD in open-world settings, where systems must detect unseen anomalies, operate without predefined taxonomies, and learn continuously. This capability is critical for safety-critical applications, enabling world models to perceive and adapt to unexpected events.

Recent open-set VAD \cite{acsintoae2022ubnormal, zhu2022towards, hirschorn2023normalizing} and open-vocabulary VAD \cite{wu2024open, li2025anomize, xu2025plovad} approaches have developed promising capabilities for detecting previously unseen anomaly types. However, single-scene training limits their generalization to unseen scenarios. Some methods \cite{zanella2024harnessing, yang2024follow, shao2025eventvad, ahn2025anyanomaly} leverage multimodal large language models (MLLMs) to obtain anomaly scores, achieving training-free detection. However, they heavily rely on frame-wise or clip-wise text outputs generated by MLLMs, which significantly limits their practical applicability.

Improving the generalization performance of VAD models faces three primary challenges: i) \textit{Limited diversity in available anomaly datasets}: Existing VAD datasets contain limited scenarios and anomaly types, which restricts model learning capabilities and makes them inadequate for real open-world applications; ii) \textit{Context-dependent semantic interpretations of anomalies}: Anomaly semantics vary according to different scenarios, while current methods lack sufficient semantic understanding, failing to comprehend unseen scenarios and novel anomaly types, and struggle to adapt detection targets dynamically; and iii) \textit{Spatiotemporal sparsity of anomalies}: Anomalies often occupy minimal temporal or spatial regions. The abundance of redundant visual information significantly increases computational cost. Moreover, detection models struggle to effectively leverage the coarse-grained (video-level) contextual semantic features generated by MLLMs, thereby overlooking localized anomalous patterns in spatiotemporal dimensions.

To address these challenges, we propose LAVIDA (\textbf{L}LM-\textbf{A}ssisted \textbf{VI}deo Anomaly \textbf{D}etection \textbf{A}pproach), an end-to-end VAD framework that leverages MLLMs and requires no real VAD data during training. To overcome the limited diversity of anomalies in existing datasets, we design an Anomaly Exposure Sampler that transforms widely accessible semantic segmentation datasets that contain diverse semantics into pseudo anomalies, thereby expanding VAD scenarios and anomaly types while eliminating the dependence of training on VAD data. To enhance semantic understanding, we employ an MLLM-integrated semantic feature extractor to capture clip-level semantic representations, utilizing MLLMs' open-world understanding to significantly improve anomaly semantic comprehension and resolve context-dependency challenges. To enable the model to focus on local anomaly patterns, considering the spatiotemporal sparsity of anomalies,  we apply a reverse-attention-based token compression method that substantially reduces irrelevant background visual information, and leverage learnable query tokens that simultaneously access clip-level context and frame-level details. At the end, we execute comprehensive anomaly detection at both frame and pixel granularities.

LAVIDA demonstrates exceptional generalization capabilities, achieving state-of-the-art performance in zero-shot detection scenarios. After training on the Anomaly Exposure datasets (external segmentation datasets), evaluations on four unseen datasets yield:  76.45\% AUC on UBnormal, 85.28\% AUC on ShanghaiTech, 82.18\% AUC on UCF-Crime (outperforming unsupervised methods), 90.62\% AP on XD-Violence (surpassing weakly-supervised methods), and 87.68\% pixel-level AUC on UCSD Ped2 (current state-of-the-art pixel-level zero-shot performance).  

In summary, our contributions are as follows: 

\begin{itemize}
\item We propose an end-to-end zero-shot VAD framework, LAVIDA, which leverages MLLMs to extract video anomaly semantic representations and enables frame/pixel-level open-world anomaly detection. 
\item We introduce an Anomaly Exposure Sampler: a training strategy that repurposes segmentation targets as pseudo-anomalies, enabling training without VAD data and improving adaptability to diverse scenarios.
\item We design a token compression method for LLM-based VAD model, which mitigates background interference and reduces computational costs for LLMs.
\item Extensive experiments show that our method achieves state-of-the-art zero-shot performance, and achieves competitive results w.r.t. unsupervised VAD methods at the frame level, and competitive zero-shot performance at the pixel level.
\end{itemize}
\section{Related Work}

\begin{figure*}[t]
\centering
\includegraphics[width=0.9\textwidth]{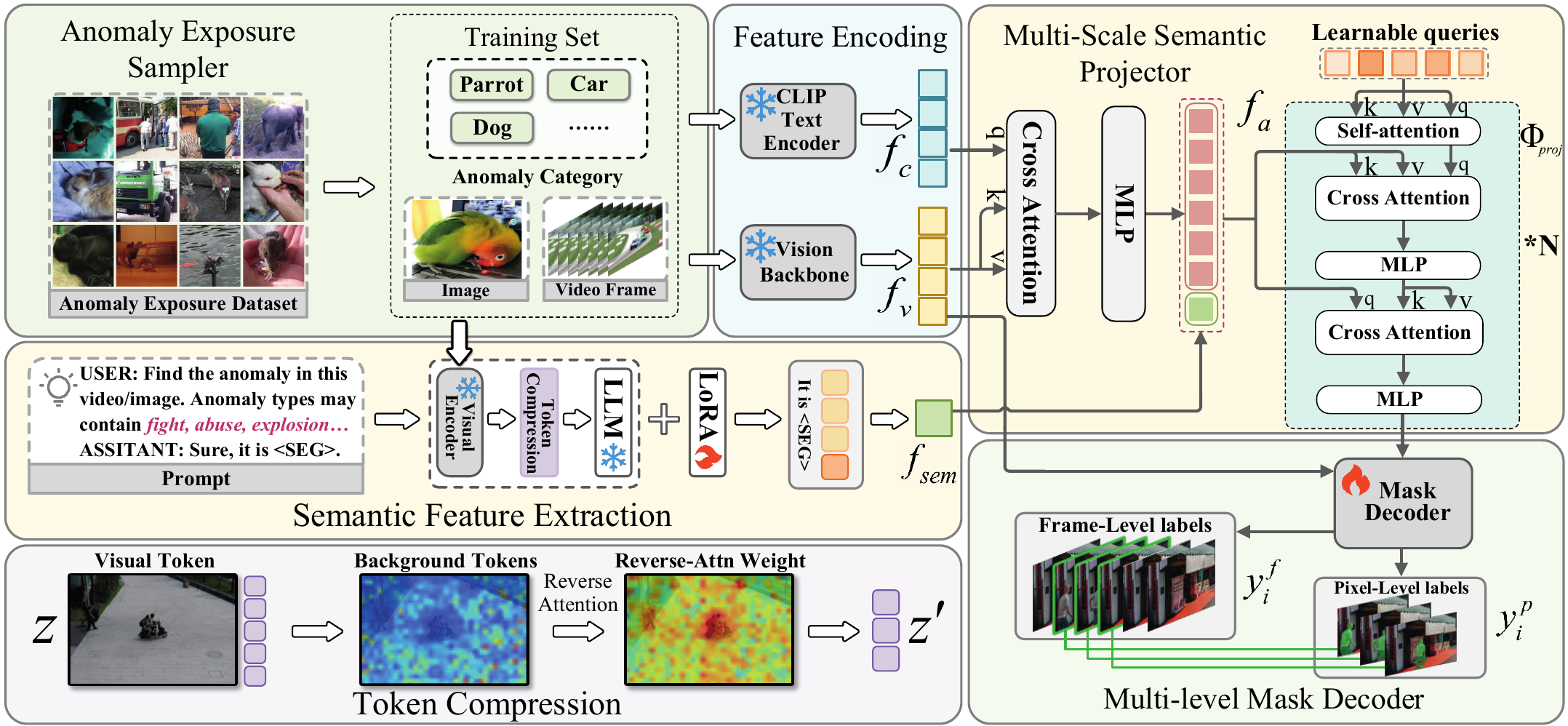}
\caption{\textbf{Overview of LAVIDA Framework.} LAVIDA is trained solely on a comprehensive Anomaly Exposure datasets, and consists of five key components: a MLLM, a text encoder, a vision backbone, a SAM2 mask decoder, and a Multi-scale Semantic Projector.}
\label{structure}
\end{figure*}

\subsection{Traditional VAD Methods}

Traditional video anomaly detection can be categorized into unsupervised and weakly-supervised approaches. Unsupervised methods assume that only normal samples exist in the training set and learn normal patterns through one-class classification (OCC) \cite{scholkopf1999support, sabokrou2018adversarially, wang2019gods, wu2019deep, zaheer2020old} or self-supervised tasks \cite{ristea2024self, yang2023video, zhong2022cascade, feng2021convolutional,hao2022spatiotemporal,liu2021hybrid}. Weakly-supervised VAD (WSVAD) detects anomalies using only video-level annotations without requiring precise temporal or spatial localization \cite{li2022self,sultani2018real,tian2021weakly,wu2021learning,zaheer2020claws}. Recent advances leverage pre-trained models and vision-language models to enhance detection performance \cite{joo2023clip, yang2024text, wu2024vadclip,  Majhi_2025_CVPR}. However, unsupervised methods struggle with unseen normal patterns and anomalous patterns similar to normal ones, and weakly-supervised methods can only recognize anomaly types present in the training set.

\subsection{Open-World VAD Methods}

To improve model generalization capabilities for unknown anomaly types, researchers have proposed open-set VAD and open-vocabulary VAD approaches. Open-set VAD was first introduced by Acsintoae et al. \cite{acsintoae2022ubnormal} with a benchmark dataset and evaluation framework. Subsequent approaches have explored evidential deep learning with normalizing flows \cite{zhu2022towards} and lightweight pose-based normalizing flows frameworks \cite{hirschorn2023normalizing}. Open vocabulary VAD enhances anomaly categorization by leveraging vision-language models. Wu et al. \cite{wu2024open} first introduce open-vocabulary video anomaly detection (OVVAD) using CLIP. Li et al. \cite{li2025anomize} leverages visual and textual information with label relations to reduce detection ambiguity. Xu et al. \cite{xu2025plovad} uses learnable prompts and graph attention networks. Nevertheless, these methods remain restricted to particular scenarios and cannot adaptively adjust detection targets based on contextual changes, which prevents these methods from achieving truly open-world VAD capabilities.

\subsection{LLM-based Video Anomaly Analysis}

Current applications of MLLMs in Video Anomaly Detection (VAD) can be categorized into training-free VAD methods and Video Anomaly Understanding (VAU) approaches. Training-free VAD methods utilize MLLMs to analyze video clips or frames, extracting anomaly scores from the generated textual outputs. Zanella et al. \cite{zanella2024harnessing} extract and refine anomaly scores from frame-wise textual outputs. Yang et al. \cite{yang2024follow} derive anomaly detection rules from training datasets for inference. Ahn et al. \cite{ahn2025anyanomaly} employ CLIP to guide MLLMs toward anomalous regions for more accurate scoring. Shao et al. \cite{shao2025eventvad} integrate dynamic graphs to mine event boundaries, enabling MLLMs to focus on event intervals. Despite eliminating training requirements, these methods rely on frame-wise or clip-wise textual outputs, incurring high temporal costs and limiting prediction granularity to frame or clip level, thus preventing spatial localization of anomalies. On the other hand, VAU methods focus on the semantic understanding capabilities of MLLMs to provide explanations for anomalies. Some studies \cite{du2024uncovering, tang2024hawk} construct interactive instruction data to deliver video-level anomaly explanations. Yuan et al. \cite{yuan2024towards} refines VAU precision to the clip level. Zhang et al. \cite{Zhang_2025_CVPR} combines detection and understanding by outputting explanations for high-probability anomalous regions. Xing et al. \cite{Xing_2025_CVPR} deploys audio data to enhance the understanding of anomalies. However, these methods prioritize using MLLMs for explanation generation while overlooking the intrinsic capability to detect unseen anomaly types.

\section{Methods}

\subsection{Preliminary}

The training dataset is an pseudo anomaly dataset $\mathcal{D}_{E} =  \{(x_i, y_i, c_i)\}_{i=1}^{N}$, where $x_i$ represents the input visual data encompassing both video and image modalities. For video samples, $v_i \in \mathbb{R}^{T \times C \times H \times W}$ where $T$, $C$, $H$, $W$ denote the number of frames, channels, height, and width, respectively, while for image samples, $I_i \in \mathbb{R}^{C \times H \times W}$. And $c_{i}=\{ c_{i,0},c_{i,1}...,c_{i,K-1}\}$ indicates the anomaly categories to be detected, and $y_i = (y_i^f, y_i^p)$ denotes the corresponding anomaly labels with frame-level label $y_i^f \in \{0,1\}^T$ and pixel-level label $y_i^p \in \{0,1\}^{T \times H \times W}$. During evaluation, the model is tested on unseen VAD datasets $\mathcal{D}_{test}  = \{(v_t, y_t, c_t)\}_{t=1}^M$, where the anomaly categories $c_t$ and video scenarios are different from the training dataset. Our objective is to predict $y_t$ in $\mathcal{D}_{test} $ under zero-shot conditions, where the test videos $v_t$ and test anomaly categories $c_t$ are not observed during training.

\subsection{Overview}

The LAVIDA framework comprises five key components, as illustrated in Fig.~\ref{structure}. First, an Anomaly Exposure Sampler reconstructs anomaly exposure dataset to form the training set. The input data then enters the Feature Encoding module, which encodes text, image, and video into feature vectors. Simultaneously, the Semantic Feature Extraction module encodes abnormal description prompts alongside vision data into a unified semantic feature via MLLM, with visual tokens being compressed by a token compression module. Thereafter, the Multi-Scale Semantic Projector fuses these semantic features with learnable query vectors and projects them into the mask decoder's latent space. Ultimately, a Multi-Level Mask Decoder decodes these latent space features to output frame-level and pixel-level anomaly scores.

\subsection{Anomaly Exposure Sampler}  

\begin{figure}[t]
\centering
\includegraphics[width=0.95\columnwidth]{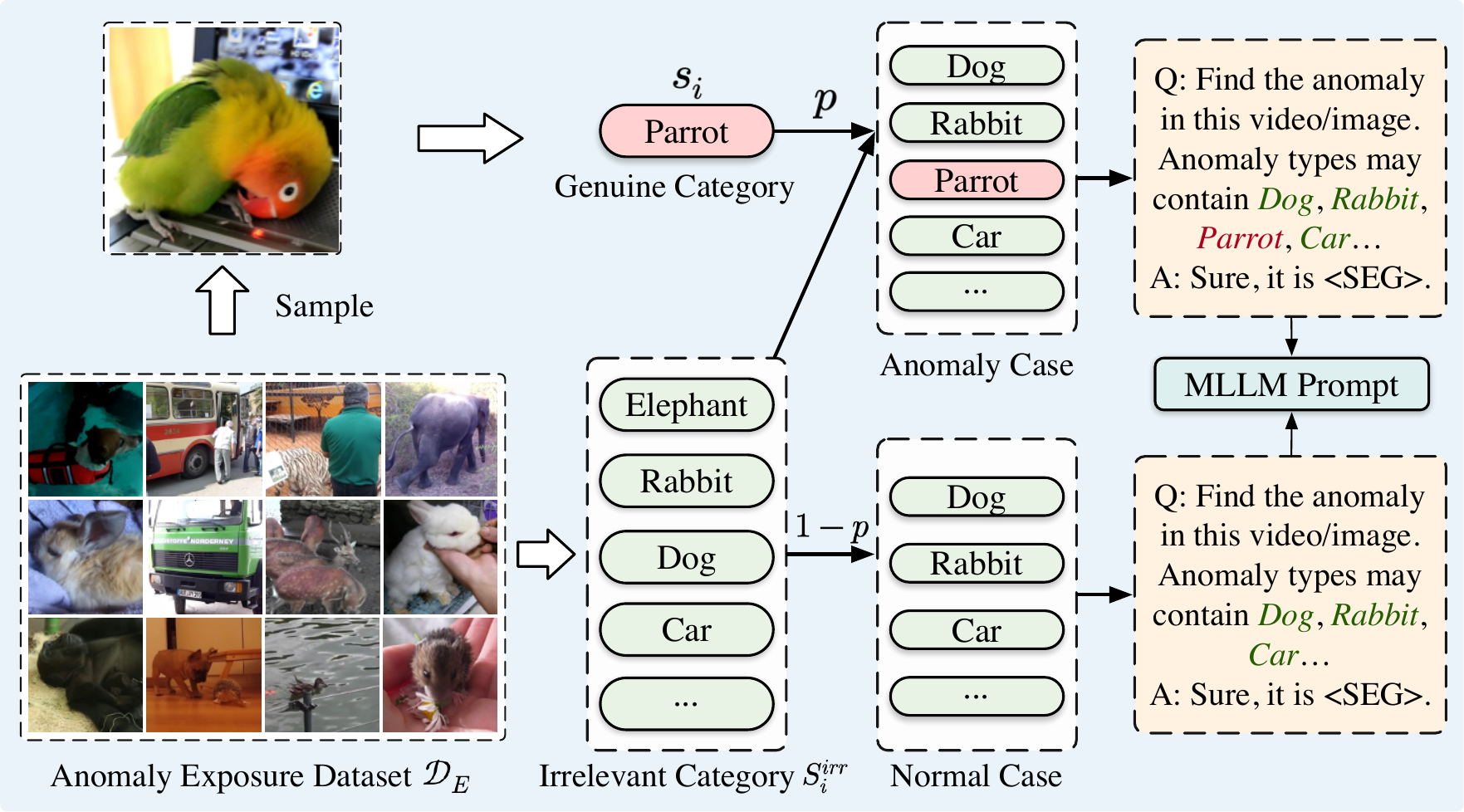} 
\caption{\textbf{Anomaly Exposure Sampler:} We sample irrelevant categories from other samples to create anomaly categories, randomly designate samples as anomalous or normal based on probability.}
\label{fig:anomaly_exposure}
\end{figure}

Visual semantic segmentation datasets provide rich scene diversity and comprehensive semantic categories. However, these datasets cannot be directly applied to VAD tasks, since anomalies occur rarely in datasets. To address this problem, we propose a two-step transformation of the anomaly exposure dataset, as illustrated in Fig.~\ref{fig:anomaly_exposure}. We define the training dataset as $\mathcal{D}_{E} =\{(x_i, y_i^p, s_i)\}_{i=1}^N$, where $s_i$ represents the text description of video clip $v_i$, and $y_i^p$ represents pixel-level category labels. Our objective is to construct $(c_i, y_i^f)$ for each sample, thereby transforming $\mathcal{D}_{E} =\{(x_i, y_i^p, s_i)\}_{i=1}^N$ into $\mathcal{D}_{E} =\{(x_i, y_i, c_i)\}_{i=1}^N$.

For anomalous samples, only sparse anomaly events exist within the video. This means that $c_i$ contains few content-relevant categories, with the majority being irrelevant. To construct $c_i$ for each sample in $\mathcal{D}_{E} $, we introduce irrelevant categories from other samples within the same dataset, thereby requiring models to distinguish genuine anomaly categories from irrelevant ones. This can be represented as follows:
\begin{equation}
S_i^{irr}=\left\{ s_{j}\,\big|\,j\sim \text{Unif} \big( \{ 1,...,n\} \setminus \{ i\} \big),\  |J|=K_{E}-1\right\}
\end{equation}
where $S_i^{irr}$ represents the set of irrelevant categories for the $i$-th sample, constructed by uniformly sampled from other samples in $\mathcal{D}_{E} $. $K_{E}$ is the total number of categories. In practice, $K_{E}$ is set as a random parameter to enable the MLLM to handle arbitrary numbers of anomaly types.

To model anomaly rarity, each sample is randomly labeled as normal (probability $1-p$) or anomaly (probability $p$). For anomalous samples, the category set $c_i$ combines genuine and irrelevant categories, with frame labels $y_i^f$ set to positive. Normal samples contain only irrelevant categories and are assigned negative labels. Such an operation is demonstrated as,
\begin{equation}
(c_i, y_i^f) = \begin{cases}
(S_i^{irr} \cup \{s_i\}, \mathbf{1}^T) & \text{with } p \\
(S_i^{irr}, \mathbf{0}^T) & \text{with } 1-p
\end{cases}
\end{equation}
where $(c_i, y_i^f)$ denotes the output category set and frame-level labels for the $i$-th sample, $T$ represents the total number of frames, and $p$ controls the anomaly sampling probability.

\subsection{Visual Token Compression}
Anomaly objects typically constitute only  a small fraction of visual data, while backgrounds constitute the vast majority. Excessive irrelevant background tokens degrade MLLM reasoning and incur substantial computational costs. We aim to deploy a training-free approach to compress background tokens while retaining anomaly-relevant features.

For VAD tasks, directly identifying anomalous tokens is difficult since the sparse spatial-temporal distribution of anomalous objects. However, background tokens are characterized by numerical predominance and high feature similarity, making them readily identifiable through density estimation. After visual encoding, the token features are represented as $Z\in \mathbb{R}^{L_z\times D_{z}}$, where $L_z$ is the number of visual tokens and $D_z$  is the token dimensionality. We compute the local density within the KNN neighborhood $N_{k}(z_{i})$ for each token $z_i$ as:
\begin{equation}
    \rho (z_{i})=\frac{k}{\sum_{z_{k}\in N_{k}(z_{i})} \| z_{i}-z_{k}\|^{2} } 
\end{equation}

We select the top-$L_r$ tokens with the highest density to form the background reference set $Z^{b} \in \mathbb{R}^{L_{r} \times D_z}$. To identify anomaly candidates, we employ a localized reverse attention mechanism \cite{huang2017semanticsegmentationreverseattention}. Specifically, each token in $Z$ is assigned to its nearest neighbor in $Z^b$ based on the minimum Euclidean distance. For each background token $Z^b_i$, reverse attention is performed exclusively over its corresponding assigned tokens to highlight features most dissimilar to the background. This process is formulated as:
\begin{equation}
Z^{\prime }_i = \text{Softmax} \left( -\frac{Z^b_i Z_{\mathcal{N}_i}^T}{\sqrt{D_z}} \right) \cdot Z_{\mathcal{N}_i}
\end{equation}
where $\mathcal{N}_i = \{ j \mid \text{arg\,min}_k \|Z_j - Z^b_k\|_2 = i \}$ denotes the set of indices of tokens in $Z$ that are closest to the $i$-th background token, and $Z^{\prime } \in \mathbb{R}^{L_r \times D_z}$ represents the aggregated anomalous features. This mechanism effectively compresses visual tokens into a compact $L_r$-length representation $Z^{\prime }$.

\subsection{Anomaly Semantics Extraction}
Existing VAD methods are constrained by limited semantic comprehension capabilities, failing to understand anomalies in unseen scenarios. To address this limitation, we leverage MLLMs following previous work LISA \cite{lai2023lisa} to extract rich anomaly semantic features that enable robust detection across diverse scenarios.

We extend the MLLM's vocabulary with a special token $<SEG>$ to extract anomaly semantic features. Given a sample $x_i$ the corresponding anomaly category $c_i$, we fill $c_i$ into several predefined templates to construct the text prompt for the MLLM. For example: \textbf{USER}: \textit{Find the anomaly in this video. Anomaly types may contain $\{c_i\}$.}  \textbf{ASSISTANT}: \textit{Sure, it is $<SEG>$.}". The $<SEG>$ token aggregates semantic information, and we extract its last-layer embedding as the anomaly semantic feature. Such an operation is demonstrated as:
\begin{equation}
    f_{sem}=\Phi_{MLLM} \left( \mathbf{x},\mathbf{c}\right)  
\end{equation}
where $\Phi_{MLLM}$ is the MLLM, $\mathbf{x}$ is the input samples, $\mathbf{c}$ is the anomaly categories that are used to construct prompts, and $f_{sem}$ is the extracted anomaly semantic feature. 

\subsection{Feature Encoding}
For the input vision data $x_i$ and anomaly categories $c_i$, we employ a vision backbone $\Phi_{v}$ to extract visual features and a CLIP text encoder $\Phi_{t}$ to extract textual features for anomaly categories:
\begin{equation}
    f_{v}=\Phi_{v} (x_i),\  \  f_{c}=\Phi_{t} (c_i)
\end{equation}
where $f_v \in \mathbb{R}^{T \times N_p \times D_v} $ represents vision feature. $N_p$ denotes the number of patches within a single frame. $f_c \in \mathbb{R}^{K \times D_t}$ represents anomaly category feature.

\begin{table*}[htbp]
\centering
\footnotesize
\fontsize{9}{12}\selectfont
\setlength{\tabcolsep}{4pt}
\begin{tabular}{l|c|c|c|c|c|c}
\hline
\hline
\multirow{2}{*}{\textbf{Methods}} & \multirow{2}{*}{\textbf{Venue}} & \multirow{2}{*}{\textbf{Training}} & \textbf{UBnormal} & \textbf{ShanghaiTech} & \textbf{UCF-Crime} & \textbf{XD-Violence} \\
 &  &  & \textbf{AUC (\%)} & \textbf{AUC (\%)} & \textbf{AUC (\%)} & \textbf{AP (\%)} \\
\hline
MemAE~\cite{gong2019memorizingnormalitydetectanomaly} & ICCV'19 & Unsupervised & - & 71.2 & - & - \\
GODS~\cite{wang2019gods} & ICCV'19 & Unsupervised & - & - & 70.4 & 61.56 \\
MSMA~\cite{mahmood2020multiscale} & ICLR'21 & Unsupervised & - & 76.7 & 64.5 & - \\
GCL~\cite{zaheer2022generative} & CVPR'22 & Unsupervised & - & 79.62 & 74.2 & - \\
FastAno~\cite{park2022fastano} & WACV'22 & Unsupervised & - & 72.2 & - & - \\
FPDM~\cite{yan2023feature} & ICCV'23 & Unsupervised & 62.7 & 78.6 & 74.7 & - \\
MULDE~\cite{micorek2024mulde} & CVPR'24 & Unsupervised & 72.8 & 81.3 & 78.5 & - \\
AED-MAE~\cite{ristea2024self} & CVPR'24 & Unsupervised & 58.5 & 79.1 & - & - \\
MA-PDM~\cite{zhou2025video} & AAAI'25 & Unsupervised & 63.4 & 79.2 & - & - \\
\hdashline
CLIP-TSA~\cite{joo2023clip} & ICIP'23 & Weakly-Supervised & - & - & 87.58 & 82.19 \\
TPWNG~\cite{yang2024text} & CVPR'24 & Weakly-Supervised & - & - & 87.79 & 83.68 \\
VadCLIP~\cite{wu2024vadclip} & AAAI'24 & Weakly-Supervised & - & - & 88.02 & 84.51 \\
Holmes-VAU~\cite{Zhang_2025_CVPR} & CVPR'25 & Weakly-Supervised & - & - & 87.68 & 88.96 \\
VERA~\cite{ye2024veraexplainablevideoanomaly} & CVPR'25 & Weakly-Supervised & - & - & 86.55 & 56.27 \\
PI-VAD~\cite{majhi2025just} & CVPR'25 & Weakly-Supervised & - & - & \textbf{90.33} & 85.37 \\
Anomize~\cite{Li_2025_CVPR}  & CVPR'25 & Weakly-Supervised & - & - & 84.49 & 69.31 \\
\hdashline
AnomalyRuler~\cite{yang2024follow} & ECCV'24 & Few-Shot & 71.9 & 85.2 & - & - \\
\hdashline
LAVAD~\cite{zanella2024harnessing} & CVPR'24 & Zero-Shot & - & - & 80.82 & 62.01 \\
AnyAnomaly~\cite{ahn2025anyanomaly} & WACV'26 & Zero-Shot & 74.5 & 79.7 & 80.7 & - \\
EventVAD~\cite{shao2025eventvad} & ACM'25 & Zero-Shot & - & - & 82.03 & 64.04 \\
\hline
\rowcolor{gray!15}\textbf{Ours} & - & Zero-Shot & \textbf{76.45} & \textbf{85.28} & 82.18 & \textbf{90.62} \\
\hline
\end{tabular}%
\caption{\textbf{Frame-level zero-shot performance compared with state-of-the-art methods.} We utilize AUC as the evaluation metric for UBnormal, ShanghaiTech and UCF-Crime datasets, and AP for the XD-Violence dataset. The best results are highlighted in bold.}
\label{tab:zero-shot}
\end{table*}

\subsection{Multi-Scale Semantic Projector}

While the MLLM effectively extracts semantic features for video anomalies, these representations remain at the video level without frame-specific granularity. To address this limitation, we propose a Multi-Scale Semantic Projector that integrates video-level semantic features with frame-level ones, generating frame-specific features $f_{proj}\in \mathbb{R}^{T\times D_{m}}$ that are projected into the mask decoder to guide fine-grained detection in each frame.

To extract frame-level local anomaly information from the video sequence, we employ cross-attention mechanisms between the anomaly category features and vision features, as demonstrated in the following formulation:
\begin{equation}
    f_{a}=\mathbf{W}_{o} \cdot \text{CrossAttn} \left( \mathbf{W}_{c} f_{c},W_{v}f_{v},W_{v}f_{v}\right)  
\end{equation}
where $f_v \in \mathbb{R}^{T \times L \times D_v}$ is vision features. $\mathbf{W}_c \in \mathbb{R}^{D_c \times D_l}$, $\mathbf{W}_v \in \mathbb{R}^{D_v \times D_l}$, and $\mathbf{W}_o \in \mathbb{R}^{D_l \times D_{a}}$ are learnable projection matrices. $D_l$ and $D_a$ represent the intermediate layer feature dimension of the output MLP and the hidden layer features of the multi-scale semantic projector, respectively. $f_a \in \mathbb{R}^{T \times K \times D_{a}}$ is the output frame-level semantic features, containing the anomaly target information for each frame. 

We expand $f_{sem}$ along the temporal dimension and apply a mapping matrix $W_{LLM}$. Then we concatenate it with $f_a$. The combined features are projected into the latent space of the mask decoder via a Q-Former-like projector. Drawing inspiration from SAM, we formulate the projector as a two-way transformer architecture, as illustrated in Fig.~\ref{structure}, to facilitate the mutual updating of both learnable queries and the extracted features $f_{sem}$ and $f_a$:
\begin{equation}
    f_{proj}=\Phi_{proj} \left( [W_{LLM}f_{sem},f_{a}]\right)  
\end{equation}
where $\Phi_{proj}$ is the projector and $f_{proj}\in \mathbb{R}^{T\times D_{m}}$ represents the projected feature, and $D_{m}$ is the latent dimension of Multi-Level Mask Decoder.

\subsection{Multi-Level Mask Decoder}
Existing VAD models typically focus on frame-level anomaly scores, limiting their detection granularity. To address this, our approach introduces a Multi-Level Mask Decoder initialized from SAM to enable both frame-level and pixel-level anomaly detection.

We feed $f_{proj}$ as the sparse prompt embedding of SAM2. After integrating the visual features $f_{v}$, the mask decoder produces pixel-level scores and object score logits. The object score logits indicate the confidence of target object presence within the image or frame, which we leverage as the frame-level anomaly score. This process can be formulated as follows:
\begin{equation}
    \hat{y}^{f}_{i}, \hat{y}^{p}_{i}  = \Phi_{d} \left( f_{proj}, f_{v}\right)  
\end{equation}
where $\hat{y}^{f}_{i}$ represents the frame-level score, $\hat{y}^{p}_{i}$ denotes the pixel-level score,  and $\Phi_{d}$ is the mask decoder of SAM2.

\subsection{Objective Function}

The objective function comprises two components: $\mathcal{L}_{txt}$ and $\mathcal{L}_{seg}$.

\begin{equation}
    \mathcal{L} = \lambda_{txt} \mathcal{L}_{txt} + \lambda_{seg} \mathcal{L}_{seg}
\end{equation}
where $\lambda_{txt}$ and $\lambda_{seg}$ are loss weight, $\mathcal{L}_{txt}$ represents the text generation loss of the MLLM, and $\mathcal{L}_{seg}$ denotes the anomaly detection loss that encompasses both frame-level and pixel-level performance enhancement. To facilitate optimization, we adopt SAM2's training loss for $\mathcal{L}_{seg}$.

\section{Experiment}
\label{sec:experiment}
Our training dataset includes a diverse collection of segmentation datasets without any VAD datasets. Detailed information regarding datasets, configurations, and additional results can be found in the supplementary material.

\begin{figure*}[t]
\centering
\includegraphics[width=0.98\textwidth]{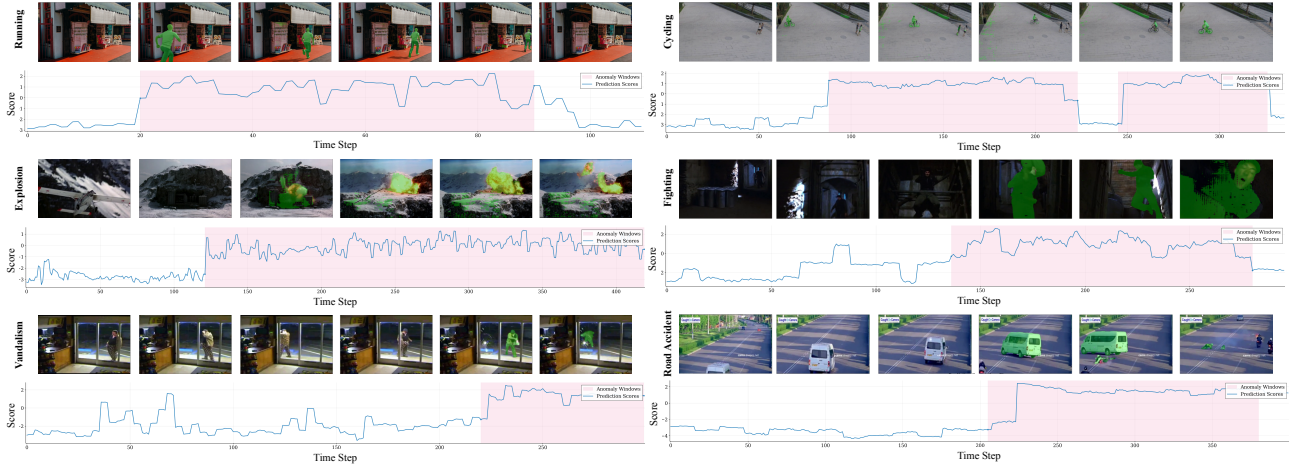} 
\caption{\textbf{Qualitative Results for Anomaly Detection.} For each case, the first row presents pixel-level detection results whitch are masked by green. The second row displays frame-level anomaly scores, with temporal intervals of anomalous events marked in pink.}
\label{quantity_exp}
\end{figure*}

\subsection{Qualitative Results}

\subsubsection{Frame-Level Zero-Shot Evaluation}
In Tab.~\ref{tab:zero-shot}, we present a comprehensive comparison of our proposed method against other SOTA approaches under zero-shot conditions across the UBnormal, ShanghaiTech, UCF-Crime, and XD-Violence datasets. The compared methods encompass four categories: unsupervised methods, weakly-supervised methods, few-shot methods, and zero-shot methods.

The experimental results demonstrate that our method attains 76.45\%, 85.28\% and 82.18\% on the UBnormal, ShanghaiTech and UCF-Crime datasets, surpassing SOTA unsupervised, zero-shot and few-shot methods. On the XD-Violence dataset, our approach achieves 90.62\%, outperforming SOTA methods. These datasets vary in both scenarios and anomaly types. This superior performance demonstrates the effectiveness of our proposed method in handling both unseen scenarios and novel anomaly categories.

Our method does not surpass weakly-supervised approaches on UCF-Crime, which we attribute to the limitations of existing MLLMs in comprehending low-resolution videos. In contrast, UBnormal, ShanghaiTech, and XD-Violence are high-resolution datasets, where small abnormal targets remain distinguishable.

\subsubsection{Pixel-Level Zero-Shot Evaluation} 
We evaluate the zero-shot pixel-level performance of our method against SOTA approaches on the UCSD Ped2 dataset, as presented in Tab.~\ref{tab:pixel}. Our method achieves a pixel-level AUC of 87.68\%, which represents a substantial improvement of 12.57\% over the current SOTA method. This significant enhancement demonstrates that our approach possesses strong zero-shot anomaly localization capability in the spatial dimension.

\begin{table}[ht]
\centering
\scriptsize
\setlength{\tabcolsep}{8pt}
\begin{tabular}{l|l|c}
\hline
\hline
\textbf{Method} & \textbf{Training} & \textbf{AUC(\%)} \\
\hline
AdaCLIP~\cite{cao2024adaclip} & Finetune & 53.06 \\
AnomalyCLIP~\cite{zhou2023anomalyclip} & Finetune & 54.25 \\
DDAD~\cite{mousakhan2024anomaly} & Supervised. & 55.87 \\
SimpleNet~\cite{liu2023simplenet} & Supervised & 52.49 \\
DRAEM~\cite{zavrtanik2021draem} & Supervised & 69.58 \\
TAO~\cite{huang2025track} & Finetune & 75.11 \\
\hline
\rowcolor{gray!15}\textbf{Proposed} & \textbf{Zero-Shot} & \textbf{87.68} \\
\hline
\end{tabular}%
\caption{\textbf{Pixel-level performance on UCSD Ped2.}}
\label{tab:pixel}
\end{table}

\begin{figure}[t]
\centering
\includegraphics[width=0.95\columnwidth]{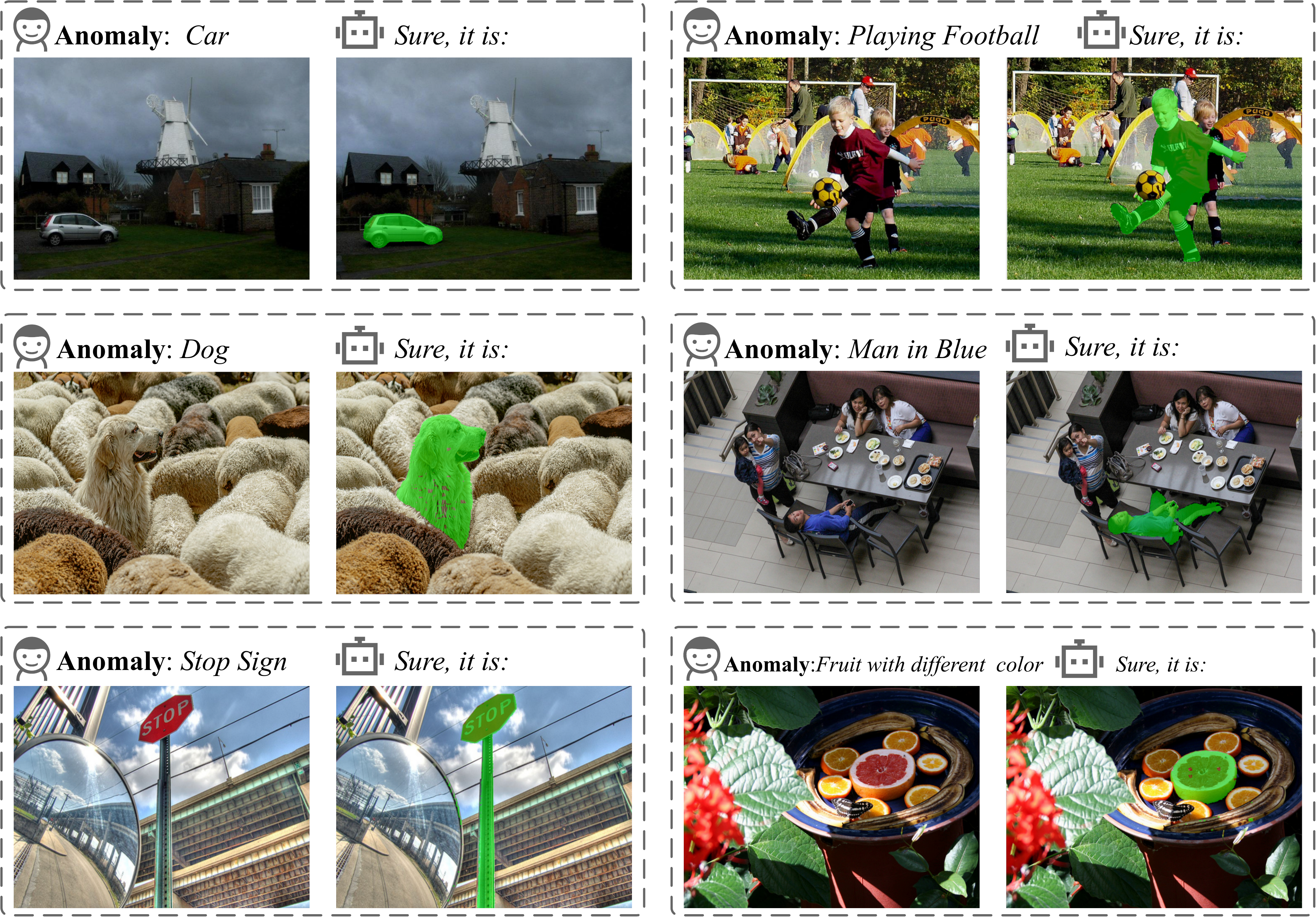} 
\caption{\textbf{Quantitative Visualizations for Open-World Scenarios. }The left panel shows the original image, and the right panel highlights detected anomalies with green masks.}
\label{fig:openworld_exp}
\end{figure}

\subsection{Quantitative Results}

Fig.~\ref{quantity_exp} presents quantitative results for anomaly detection across six representative cases from different VAD datasets. For each case, the lower row displays frame-level anomaly scores over time with anomalous intervals highlighted in pink, and the upper row shows pixel-level detection results at corresponding time steps with anomaly targets masked in green. Higher scores in pink regions and lower scores elsewhere demonstrate effective frame-level detection. Results show our model accurately identifies anomalous frames in unseen scenarios and generates precise pixel-level scores that delineate anomaly boundaries.

Fig.~\ref{fig:openworld_exp} presents open-world detection results. For each case, the target anomaly category is specified in the text. The left panel shows original images, while the right panel presents detection results with anomalies highlighted in green. Our method demonstrates robust performance and strong reasoning capabilities in identifying arbitrary anomaly types across diverse scenarios.

\subsection{Ablation Studies}

\subsubsection{Analysis of the Anomaly Exposure Sampler}

\begin{table}[htbp]
\centering
\scriptsize
\setlength{\tabcolsep}{3pt}
\begin{tabular}{c|c|c|c}
\hline
\multirow{2}{*}{$\max(K_E)$} & \textbf{ShanghaiTech} & \textbf{UCF-Crime} & \textbf{XD-Violence} \\
 & \textbf{AUC (\%)} & \textbf{AUC (\%)} & \textbf{AP (\%)} \\ 
\hline
10 & 73.34 & 67.11 & 55.44  \\ 
20 & 80.13 & 80.02 & 91.20   \\ 
30 & 85.02 & 82.06 & 90.30   \\ 
40 & 76.41 & 77.93 & 75.32   \\ 
\hline
\end{tabular}
\caption{\textbf{Effect of the number of anomaly categories introduced in the anomaly exposure dataset.}}
\label{table:anomaly_categories} 
\end{table}

Tab.~\ref{table:anomaly_categories} shows the effect of anomaly category count $K_E$ in the anomaly exposure dataset. To enable MLLMs to comprehend arbitrary anomaly types, we set $K_E$ as a random variable and evaluate the impact by controlling $\max(K_E)$. Results show that performance is poor when $\max(K_E)=10$, improves as the value increases, and reaches optimum around 30. Further increases beyond 30 cause performance degradation, which we attribute to excessively lengthy prompts that reduce the model's focus on individual anomaly types.

\subsubsection{Analysis of Token Compression}

\begin{figure}[t]
\centering
\includegraphics[width=0.9\columnwidth]{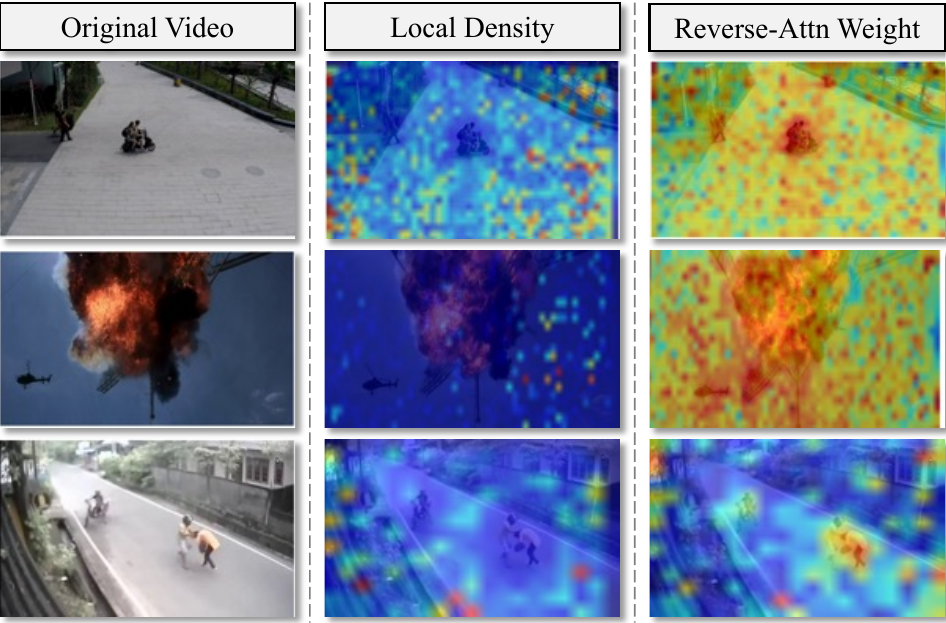} 
\caption{\textbf{Token Compression Process}. The first column shows the original video frames. The second column shows local density, and the third shows reverse attention weights. Warmer colors indicate higher values.}
\label{fig:attn_weight}
\end{figure}

\begin{figure}[htbp]
    \centering
    \begin{subfigure}[b]{0.8\columnwidth}
        \centering
        \includegraphics[width=\columnwidth]{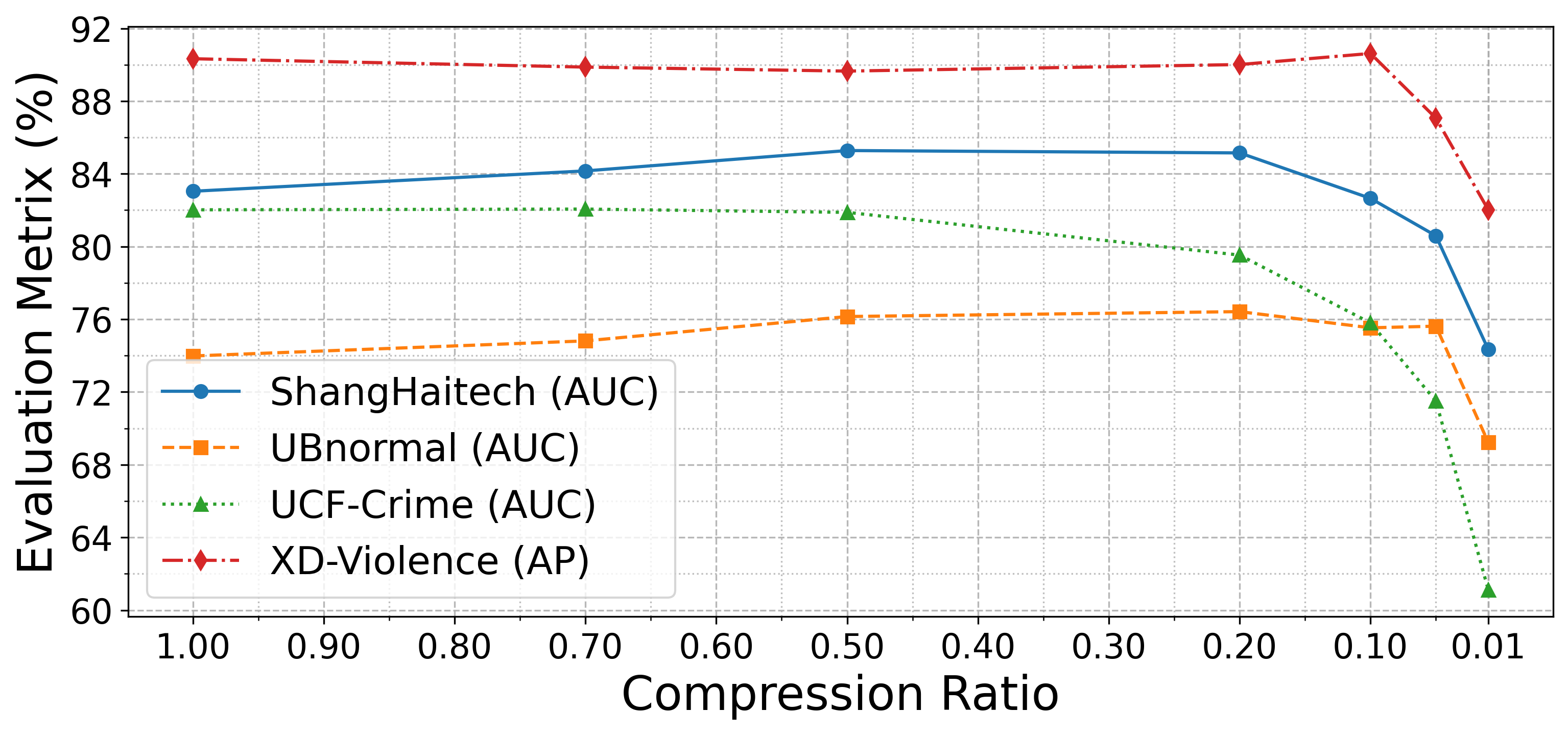}
        \caption{Compression ratio vs. model performance.}
        \label{fig:score_ratio}
    \end{subfigure}
    \hfill
    \begin{subfigure}[b]{0.8\columnwidth}
        \centering
        \includegraphics[width=\columnwidth]{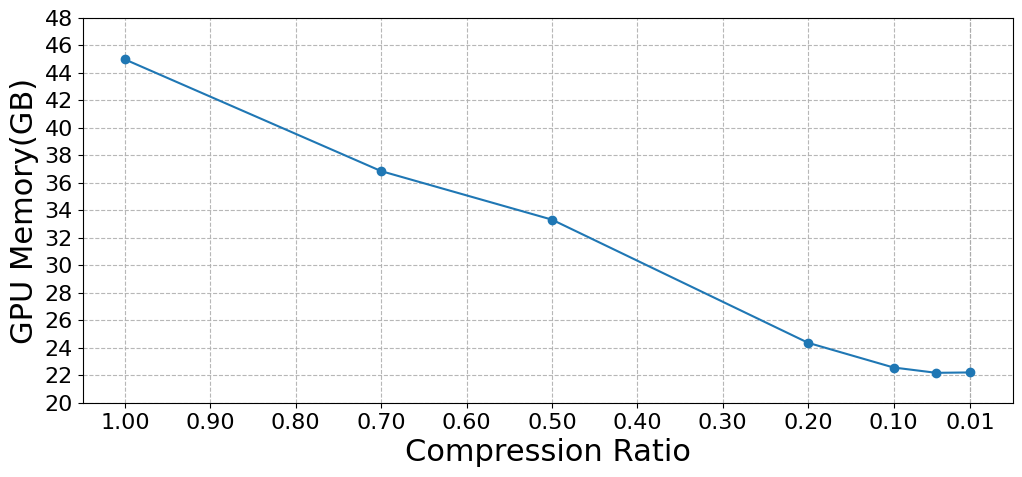}
        \caption{Compression ratio vs. GPU memory usage.}
        \label{fig:gpu_ratio}
    \end{subfigure}
    \caption{\textbf{Impression of compression ratio.} We normalize the frame size across all datasets to ensure a consistent number of total visual tokens. GPU memory usage is recorded during inference stage.}
    \label{fig:both_images}
\end{figure}

Fig.~\ref{fig:attn_weight} illustrates the token compression process. We leverage local density to identify tokens with the highest density values, which correspond closely to the background regions. During the reverse attention stage, tokens whose features are highly dissimilar to the background tokens are aggregated via reverse attention weight. As shown in the third column of Fig.~\ref{fig:attn_weight}, the reverse attention weights concentrate on regions that exhibit substantial dissimilarity from the background, which are typically anomaly-prone areas.

Fig.~\ref{fig:score_ratio} presents the variation of performance across datasets with respect to the token compression ratio. For compression ratios above 0.1, performance remains relatively stable, demonstrating effective background token reduction without compromising model capability. On high-resolution datasets (UBnormal, ShanghaiTech, XD-Violence), we observe slight improvements, as background suppression enables better focus on small anomaly targets. When the compression ratio drops below 0.1, a marked performance degradation occurs due to substantial information loss caused by excessive compression. UCF-Crime's resolution is much lower than other datasets, making it the most susceptible to the visual compression.

Fig.~\ref{fig:gpu_ratio} shows the corresponding GPU memory utilization versus the compression ratio. As the number of input visual tokens decreases, GPU memory consumption exhibits a linear reduction. At a compression ratio of 0.2, GPU memory usage is reduced to 54.1\% of the baseline, with no substantial performance loss (UBNormal: +2.44\%, ShanghaiTech: +2.24\%, XD-Violence: -0.32\%, UCF-Crime: -2.49\%, Average: +0.47\%).

\subsubsection{Analysis of the Multi-Scale Semantic Projector} 

\begin{table}[t]
\centering
\scriptsize
\setlength{\tabcolsep}{3pt}
\begin{tabular}{l|c|c|c|c}
\hline
 & \multicolumn{3}{c|}{\textbf{frame-level}} & \textbf{pixel-level} \\
\cline{2-5}
\multirow{2}{*}{\textbf{Adapter}} & \textbf{ShanghaiTech} & \textbf{UCF-Crime} & \textbf{XD-Violence} & \textbf{Ped2} \\
 & \textbf{AUC (\%)} & \textbf{AUC (\%)} & \textbf{AP (\%)} & \textbf{AUC (\%)} \\
\hline
MLP & 81.99 & 79.90 & 86.09 & 77.09\\
Q-Former & 82.63 & 75.54 & 83.93 & 71.85\\
Proposed & 85.28 & 82.06 & 90.62 & 87.68\\
\hline
\end{tabular}
\caption{\textbf{Comparison among different adapters.}}
\label{tab:adapter}
\end{table}

\begin{table}[tp]
\centering
\scriptsize
\setlength{\tabcolsep}{3pt}
\begin{tabular}{c|c|c|c}
\hline
\multirow{2}{*}{number of queries} & \textbf{ShanghaiTech} & \textbf{UCF-Crime} & \textbf{XD-Violence} \\
 & \textbf{AUC (\%)} & \textbf{AUC (\%)} & \textbf{AP (\%)} \\ 
\hline
24 & 58.85 & 66.63 & 50.14  \\ 
32 & 77.41 & 77.96 & 88.31   \\ 
48 & 80.64 & 80.06 & 90.30   \\ 
64 & 77.96 & 67.11 & 89.15   \\ 
\hline
\end{tabular}
\caption{\textbf{Effect of the number of learnable queries.}}
\label{table:num_queries}
\end{table}

To validate the effectiveness of our Multi-Scale Semantic Projector, we compared it against MLP and Q-Former at both frame-level and pixel-level. The experimental results are presented in Tab.~\ref{tab:adapter}. The improvements at frame-level demonstrate the capability to capture temporal anomaly cues, while the enhancements at pixel-level indicate the ability to detect spatially sparse anomalies.

Tab.~\ref{table:num_queries} presents the effect of learnable query count on zero-shot detection performance. With 24 queries, the model achieves suboptimal results due to limited representational capacity. Performance improves as query count increases, reaching a peak before declining when queries become excessive, causing convergence difficulties.
\vspace{0.2cm}
\section{Conclusion}

In this paper, we propose LAVIDA, an end-to-end zero-shot VAD approach that leverages MLLM and token compression algorithm to extract semantic anomaly features and an anomaly exposure sampler to enable anomaly detection in open-world scenarios without training VAD data. Multi-scale semantic projector is employed to extract hierarchical cues for joint frame- and pixel-level prediction. Extensive experiments the effectiveness across multiple benchmarks. We hope our work inspires further researches in developing open-world video anomaly detection and understanding.

\paragraph*{Acknowledgement}
This work is supported in part by the National Natural Science Foundation of China (No. 62272057) and the Beijing Key Laboratory of Multimodal Data Intelligent Perception and Governance.

{
    \small
    \bibliographystyle{ieeenat_fullname}
    \bibliography{main}
}

\end{document}